\documentclass{article}

\usepackage[preprint,nonatbib]{neurips_2020}

\usepackage[utf8]{inputenc}
\usepackage[T1]{fontenc}
\usepackage[colorlinks=true]{hyperref}
\usepackage{url}
\usepackage{booktabs}
\usepackage{amsfonts}
\usepackage{nicefrac}
\usepackage{microtype}
\usepackage{enumerate}
\usepackage{amsmath, amssymb}
\usepackage{color}
\usepackage{xcolor}
\usepackage{colortbl}
\usepackage{mathptmx}
\usepackage{multicol}
\usepackage{graphicx}
\usepackage{tabularx}
\usepackage{graphbox} 

\title{High-resolution land cover change from low-resolution labels: Simple baselines for the 2021 IEEE GRSS Data Fusion Contest}

\author{%
  Kolya Malkin \\
  Yale University \\
  \And
  Caleb Robinson\thanks{\texttt{caleb.robinson@microsoft.com}} \\
  Microsoft AI for Good \\
  \And
  Nebojsa Jojic \\
  Microsoft Research \\
}

\begin{document}

\maketitle

\begin{abstract}
    We present simple algorithms for land cover change detection in the 2021 IEEE GRSS Data Fusion Contest~\cite{dfc2021}. The task of the contest is to create high-resolution (1m / pixel) land cover change maps of a study area in Maryland, USA, given multi-resolution imagery and label data. We study several baseline models for this task and discuss directions for further research.
    
    See \url{https://dfc2021.blob.core.windows.net/competition-data/dfc2021_index.txt} for the data and  \url{https://github.com/calebrob6/dfc2021-msd-baseline} for an implementation of these baselines.
\end{abstract}

\section{Introduction}

We describe a possible road map towards estimating high-resolution (e.g., 1m) land cover change, using high-resolution multitemporal input imagery and low-resolution (e.g., 30m) noisy land cover labels for one or more time points. The examples here use the data from the multitemporal semantic change detection track of the 2021 IEEE Geoscience and Remote Sensing Society's Data Fusion Contest (DFC-MSD) \cite{dfc2021}. The input imagery is from the National Agriculture Imagery Program (NAIP) for the years 2013 and 2017, captured at 1m resolution, and the available labels are from the 30m-resolution National Land Cover Database (NLCD) for the years 2013 and 2016. The NLCD labels are derived from Landsat imagery, and the contest also provides multiyear Landsat data for the period between 2013 and 2017. All the data is limited to Maryland, USA.

The task exemplifies the situation commonly found worldwide. New imagery comes in faster than high-quality high-resolution labels are being created. However, some older, noisy, and low-resolution labels are often available, e.g., 30m NLCD in the United States, or 500m MODIS land cover available worldwide. How could one use machine learning to build models that predict high-resolution change without a large set of high-resolution change examples? The hope springs from the success of weakly supervised segmentation and label-super resolution research, e.g. \cite{lsr,epitomes}, which demonstrated that it is possible to train high-resolution label predictors for high-resolution input imagery using regional supervision, where a large block of land (e.g. 30m$\times$30m) is labeled with a single class, which often designates an area where land cover follows particular mixing proportions. Weak supervision was the topic of the 2020 IEEE GRSS Data Fusion Contest \cite{dfc2020}.

Taking the next step to predict land cover \emph{change} could be as straightforward as applying these techniques separately for different time points and comparing the super-resolved labels to estimate change, and in this paper we evaluate what such simple baselines could accomplish. We demonstrate some simple lessons in terms of scope of training and model selection that could aid in development of techniques specifically built for high-resolution \emph{change} detection. Such techniques would presumably not use the multiyear images in isolation, but analyze them jointly to take advantage of the fact that most land cover \emph{does not} change from year to year.

The paper is organized as follows. We first describe the data in the DFC-MSD task and illustrate the types of changes that are being scored in the contest. Then we study training neural networks to approximate functions $y=f(x)$ where $x$ are high-resolution (1m) image patches and $y$ are upsampled $30$m labels, and discuss how the choices for model complexity and scope of training data can affect the results. In particular, a highly expressive and well-trained function $f$ could learn to predict blurry labels, because the targets $y$ cone in $30\times30$ blocks, while very simple color and texture models, e.g., \cite{epitomes,contest2020}, tend to super-resolve them at 1m, as they are unable to assign the same label to $30 \times 30$ blocks of pixels with varying color and texture. Finally, we discuss possible approaches that could use all the data in concert to build more accurate prediction models. 

All data can be downloaded from the addresses listed at \url{https://dfc2021.blob.core.windows.net/competition-data/dfc2021_index.txt}; alternative fast download instructions can be found on the DFC-MSD webpage \cite{dfc2021}. Example code is available in the accompanying code repository at \url{https://github.com/calebrob6/dfc2021-msd-baseline}.

\section{The change detection task}

\subsection{Data}

The input data in the DFC-MSD comprises 9 layers covering the study area of the state of Maryland in the United States ($\sim$35,000~km$^2$). All layers are upsampled from their native resolutions to 1m / pixel and provided as 2250 aligned \emph{tiles} of dimensions not exceeding $4000\times4000$. The layers are the following:

\begin{enumerate}[(1)]
    \item \textbf{NAIP} (2 layers): 1m-resolution 4-band (red, green, blue, and near-infrared) aerial imagery from the US Department of Agriculture's National Agriculture Imagery Program (NAIP) from two points in time: 2013 and 2017.
    \item \textbf{Landsat} (5 layers): 30m-resolution 9-band imagery from the Landsat-8 satellite from five time points: 2013, 2014, 2015, 2016, and 2017. Each of these images is a median-composite from all cloud and cloud-shadow masked surface-reflectance scenes intersecting Maryland.
    \item \textbf{NLCD} (2 layers): 30m-resolution coarse land cover labels from the US Geological Survey's National Land Cover Database \cite{nlcd}, in 15 classes (see Table~\ref{tab:nlcd_to_target}) from two time points: 2013 and 2016. These labels were created in a semi-automatic way, with Landsat imagery as the principal input.
\end{enumerate}

\begin{figure}[t]
    \centering
    \includegraphics[width=0.48\textwidth]{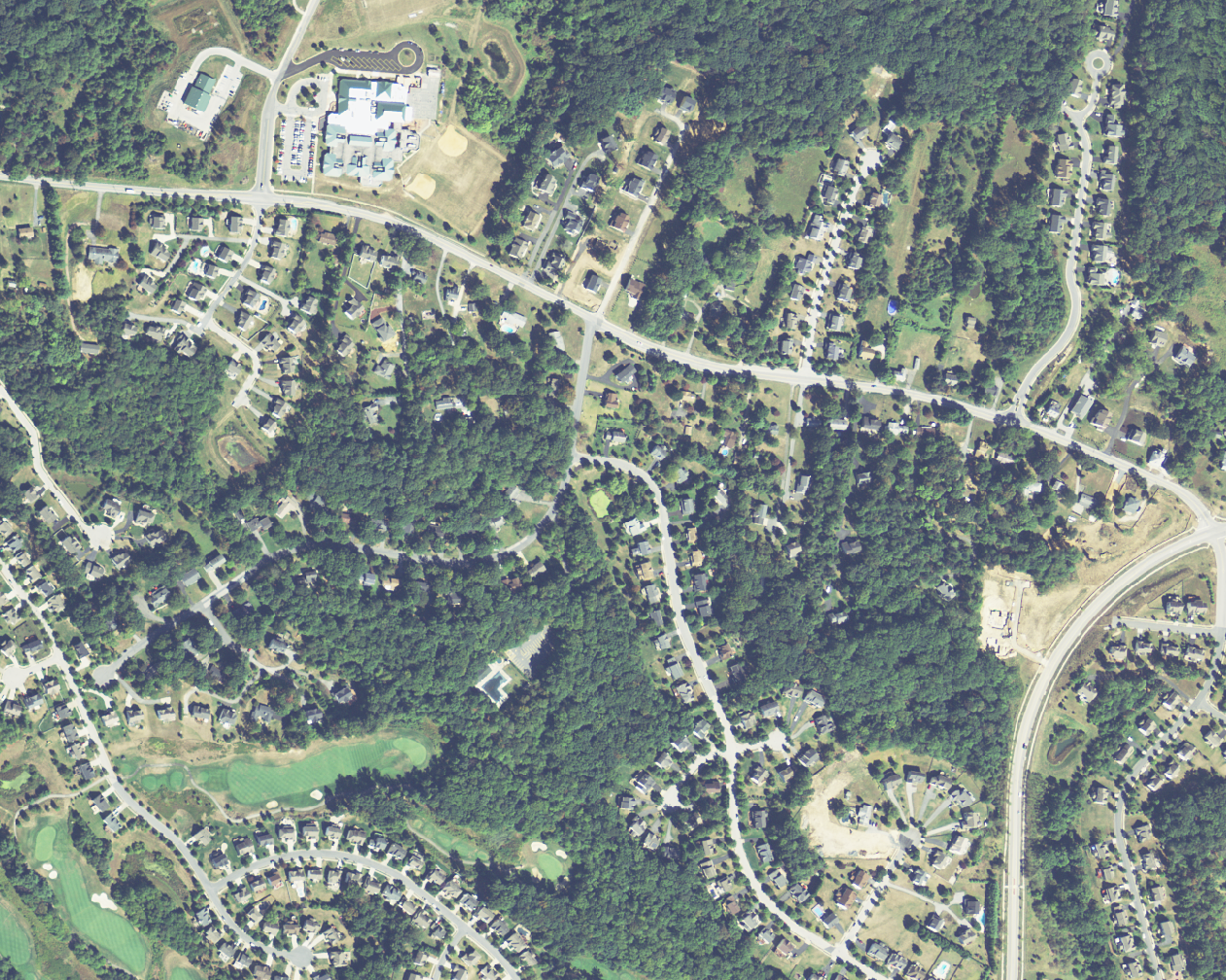}
    \includegraphics[width=0.48\textwidth]{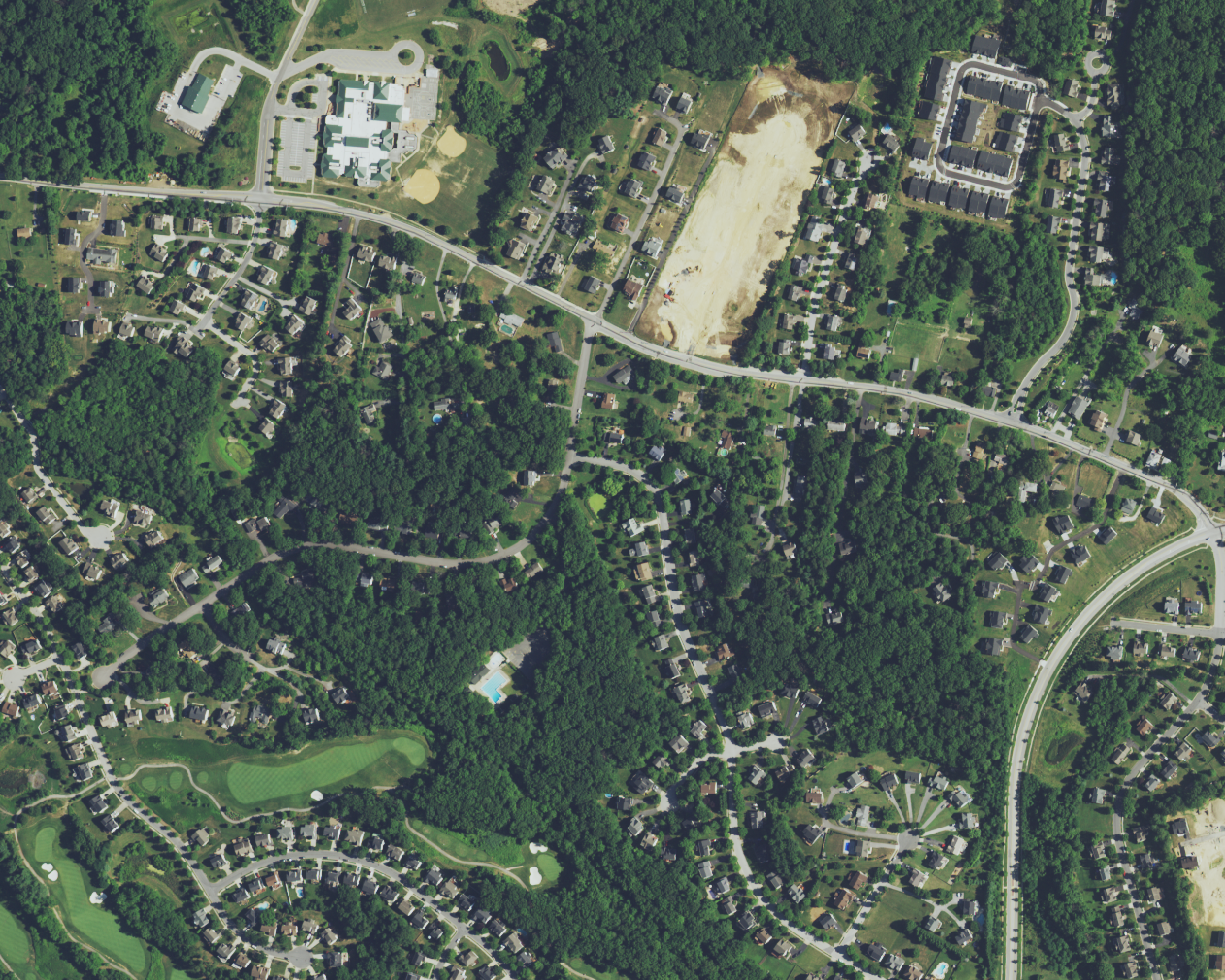}\\
    \includegraphics[width=0.48\textwidth]{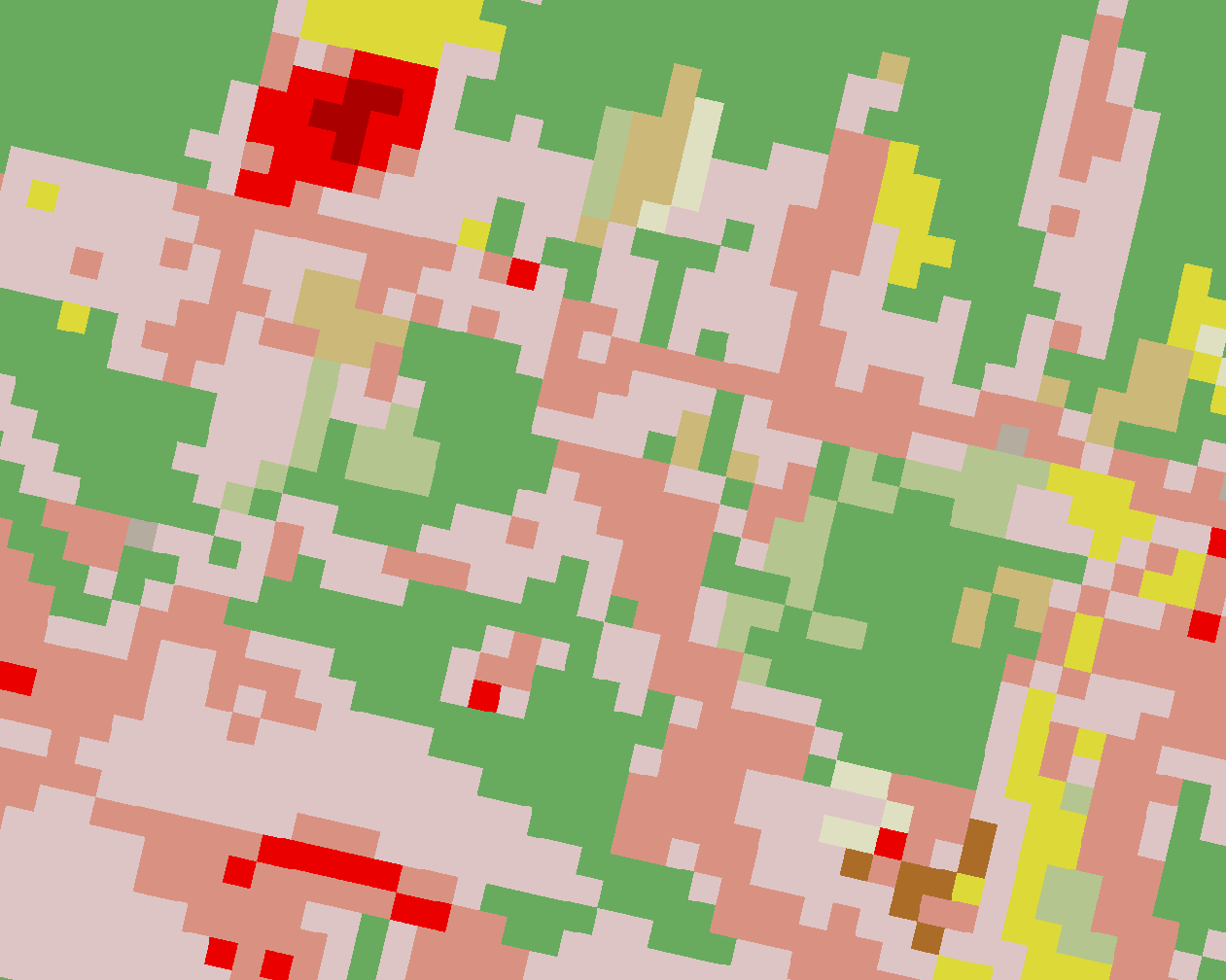}
    \includegraphics[width=0.48\textwidth]{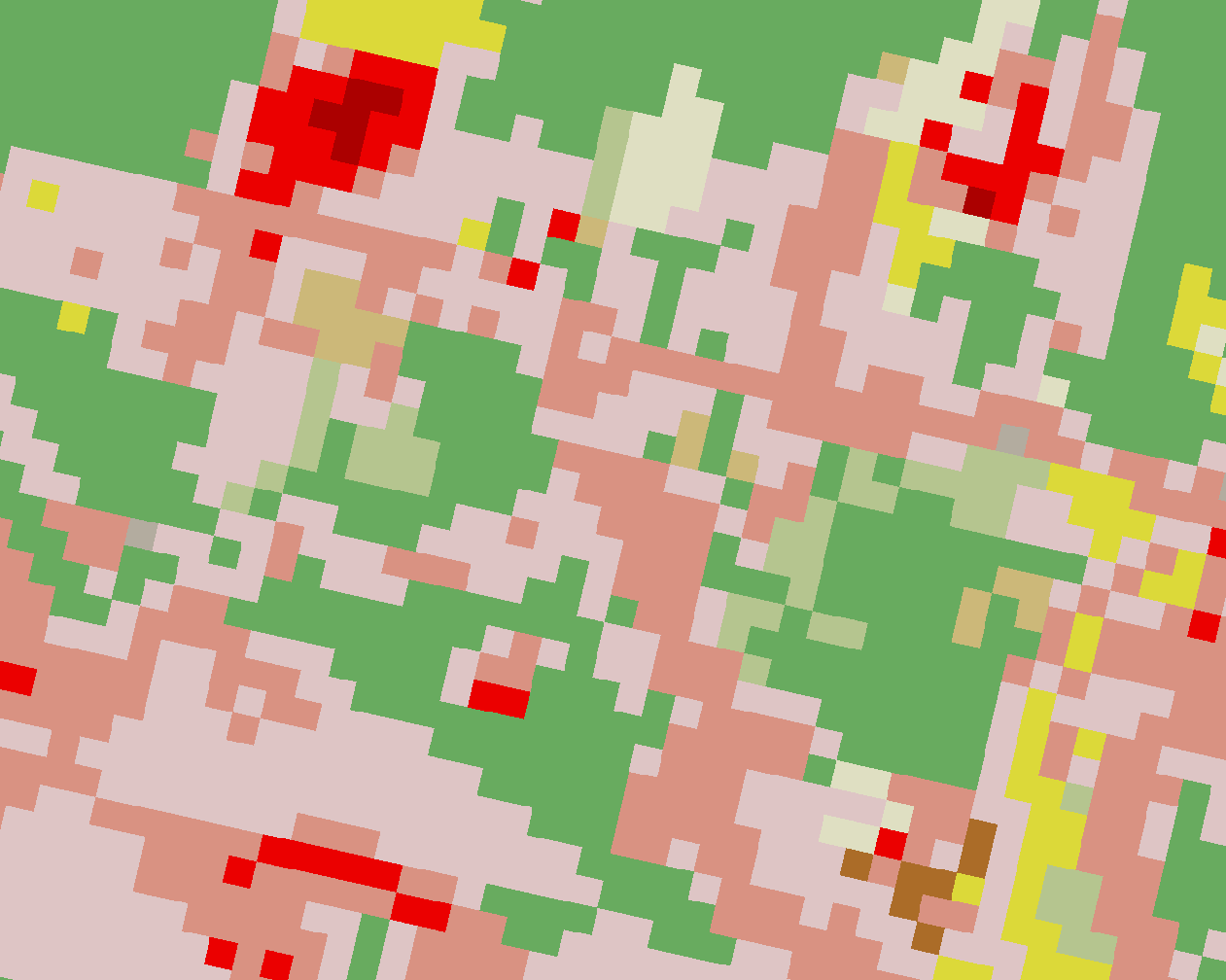}
    \caption{Four of the input data layers in the DFC-MSD. \textbf{Above: } NAIP imagery from 2013 and 2017. \textbf{Below: } Coarse NLCD land cover labels from 2013 and 2016. See Table~\ref{tab:nlcd_to_target} for the color scheme.}
    \label{fig:data_examples}
\end{figure}

An example of the input layers is shown in Figure~\ref{fig:data_examples}.

\subsection{Task}

The goal of the DFC-MSD is to classify pixels in the study area as belonging to classes describing land cover change between the times when NAIP 2013 and 2017 aerial photographs were taken. Specifically, we wish to detect the \textbf{loss} or \textbf{gain} of four land cover classes in a simplified scheme based on that of the NLCD: \textbf{water}, \textbf{tree canopy}, \textbf{low vegetation}, and \textbf{impervious surfaces}, as well as the absence of land cover change. An approximate correspondence between NLCD classes and the target classes is shown in Table~\ref{tab:nlcd_to_target}.

\definecolor{6274a4}{HTML}{466b9f}
\definecolor{e2d2d2}{HTML}{dec5c5}
\definecolor{d0a18f}{HTML}{d99282}
\definecolor{db3318}{HTML}{eb0000}
\definecolor{9b200e}{HTML}{ab0000}
\definecolor{d0cec1}{HTML}{b3ac9f}
\definecolor{9ac786}{HTML}{68ab5f}
\definecolor{558053}{HTML}{1c5f2c}
\definecolor{d9e7b4}{HTML}{b5c58f}
\definecolor{d6ca96}{HTML}{ccb879}
\definecolor{f6eab0}{HTML}{dfdfc2}
\definecolor{f8f876}{HTML}{dcd939}
\definecolor{bb9451}{HTML}{ab6c28}
\definecolor{d2e5f6}{HTML}{b8d9eb}
\definecolor{83b1d2}{HTML}{6c9fb8}

\definecolor{waterloss}{HTML}{c0c0ff}
\definecolor{treeloss}{HTML}{60c060}
\definecolor{fieldloss}{HTML}{c0ffc0}
\definecolor{imploss}{HTML}{c0c0c0}
\definecolor{watergain}{HTML}{0000ff}
\definecolor{treegain}{HTML}{008000}
\definecolor{fieldgain}{HTML}{80ff80}
\definecolor{impgain}{HTML}{808080}

\begin{table*}[t]
    \centering
    \begin{tabular}{lllllrrrr}
    \toprule
    &&&&&\multicolumn{4}{c}{Approximate class freq.}\\
& NLCD class &&& Target class & W\% & TC\% & LV\% & I\% \\\midrule
\cellcolor{6274a4}\ \ 
&Open Water
&\cellcolor{waterloss}\ \ &\cellcolor{watergain}\ \ &water
& 98
& 2
& 0
& 0
\\
 
\cellcolor{e2d2d2}\ \ 
&Developed, Open Space
&&& --
& 0
& 39
& 49
& 12
\\
 
\cellcolor{d0a18f}\ \ 
&Developed, Low Intensity
&&& --
& 0
& 31
& 34
& 35
\\
 
\cellcolor{db3318}\ \ 
&Developed, Medium Intensity
&\cellcolor{imploss}\ \ &\cellcolor{impgain}\ \ &impervious
& 1
& 13
& 22
& 64
\\
 
\cellcolor{9b200e}\ \ 
&Developed High Intensity
&\cellcolor{imploss}\ \ &\cellcolor{impgain}\ \ &impervious
& 0
& 3
& 7
& 90
\\
 
\cellcolor{d0cec1}\ \ 
&Barren Land (Rock/Sand/Clay)
&&& --
& 5
& 13
& 43
& 40
\\
 
\cellcolor{9ac786}\ \ 
&Deciduous Forest
&\cellcolor{treeloss}\ \ &\cellcolor{treegain}\ \ &tree canopy
& 0
& 93
& 5
& 0
\\
 
\cellcolor{558053}\ \ 
&Evergreen Forest
&\cellcolor{treeloss}\ \ &\cellcolor{treegain}\ \ &tree canopy
& 0
& 95
& 4
& 0
\\
 
\cellcolor{d9e7b4}\ \ 
&Mixed Forest
&\cellcolor{treeloss}\ \ &\cellcolor{treegain}\ \ &tree canopy
& 0
& 92
& 7
& 0
\\
 
\cellcolor{d6ca96}\ \ 
&Shrub/Scrub
&\cellcolor{treeloss}\ \ &\cellcolor{treegain}\ \ &tree canopy
& 0
& 58
& 38
& 4
\\
 
\cellcolor{f6eab0}\ \ 
&Grassland/Herbaceous
&\cellcolor{fieldloss}\ \ &\cellcolor{fieldgain}\ \ &low vegetation
& 1
& 23
& 54
& 22
\\
 
\cellcolor{f8f876}\ \ 
&Pasture/Hay
&\cellcolor{fieldloss}\ \ &\cellcolor{fieldgain}\ \ &low vegetation
& 0
& 12
& 83
& 3
\\
 
\cellcolor{bb9451}\ \ 
&Cultivated Crops
&\cellcolor{fieldloss}\ \ &\cellcolor{fieldgain}\ \ &low vegetation
& 0
& 5
& 92
& 1
\\
 
\cellcolor{d2e5f6}\ \ 
&Woody Wetlands
&\cellcolor{treeloss}\ \ &\cellcolor{treegain}\ \ &tree canopy
& 0
& 94
& 5
& 0
\\
 
\cellcolor{83b1d2}\ \ 
&Emergent Herbaceous Wetlands
&\cellcolor{treeloss}\ \ &\cellcolor{treegain}\ \ &tree canopy
& 8
& 86
& 5
& 0
\\
    \bottomrule
    \end{tabular}
    \caption{The approximate correspondence between the NLCD classes and the four target classes. The last four columns show \emph{estimated} frequencies of target classes of pixels labeled with each NLCD class, computed from a high-resolution land cover product, specific to Maryland, in a similar class scheme \cite{chesapeake}. The colors next to NLCD classes are those used in Figure~\ref{fig:data_examples}. The two colors next to the target classes are those used to show loss and gain respectively in Figures~\ref{fig:gt_examples} and \ref{fig:baseline_examples}.}
    \label{tab:nlcd_to_target}
\end{table*}

\begin{figure}[t]
    \centering
    \includegraphics[width=0.9\textwidth]{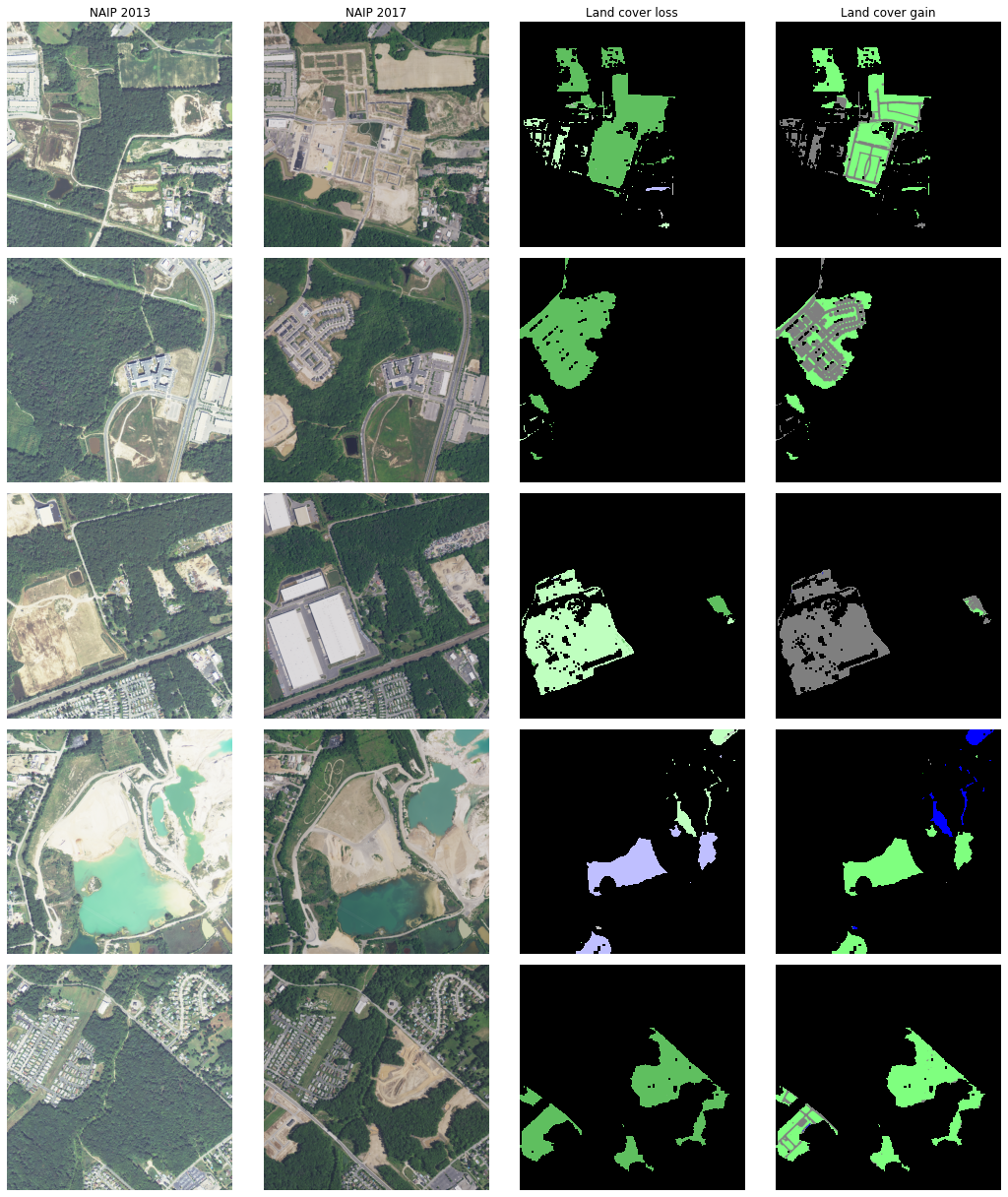}
    \caption{Examples of paired NAIP 2013 and NAIP 2017 imagery and the desired predictions. See Table~\ref{tab:nlcd_to_target} for the color scheme. Pixels with no change are shown in black.}
    \label{fig:gt_examples}
\end{figure}

We require that any pixel classified as a \textbf{gain} of one land cover class must be simultaneously classified as a \textbf{loss} of another class. Thus, the task is equivalent to creating two aligned land cover maps in the four target classes for the two time points and computing their disagreement: pixels assigned class $c_1$ in 2013 and $c_2$ in 2017 are considered \textbf{loss} of $c_1$ and \textbf{gain} of $c_2$; pixels in which the maps agree are considered \textbf{no change}. See Figure~\ref{fig:gt_examples} for examples of the desired outputs. For example, in the second row, a part of the forest is cut (loss of \textbf{tree canopy}). In its place we see some gain in \textbf{impervious surfaces} (buildings and roads), as well as some cleared fields, which are placed in the \textbf{low vegetation} category. In a few places, new trees were planted, resulting in no change in land cover between the two time points. In the first row, where forest is also cut, most of it is still undeveloped and consists of cleared fields. The undisclosed evaluation data consists of many small areas of interest such as the ones in the figure, as well as a set of areas where there has been no change. Ambiguous areas are blocked from evaluation. These include shadows and areas which humans have difficulty recognizing.

This problem is made difficult by the very different spectral characteristics of the two NAIP layers (due to differences in sensors, seasons, and time of day when they were created), as well as the lack of high-resolution label data: the only available training labels are the NLCD layers, which are coarse, created for different points in time than NAIP, and have a different class scheme.

\subsection{Contest description}

The DFC-MSD is organized in two phases: the \emph{validation} phase, in which contestants have 100 attempts to submit their proposed land cover change maps for a subset of the study area to an evaluation server and receive feedback on their performance, and the \emph{test} phase, in which contestants have 10 attempts to submit change maps for a different subset of the study area.

For the validation phase, a set of 50 out of the 2250 tiles were selected for evaluation. Within these tiles, regions of interest were identified and high-resolution change maps created by the contest organizers. Some regions of interest contain pixels with land cover change, while others have only the \textbf{no change} label. While participants submit predictions for all 50 tiles, evaluation is performed only over the areas of interest; predictions for pixels outside areas of interest are ignored. The class statistics for the validation set are shown in Table~\ref{tab:class_stats}.

The scoring metric is the average intersection over union (IoU) between the predicted and ground truth labels for eight classes (loss and gain of each of the four target classes, excluding \textbf{no change}). (Recall that the IoU for a class $c$ is defined as 
\begin{equation}
    \frac{\text{\# pixels labeled $c$ in prediction \emph{and} in ground truth}}{\text{\# pixels labeled $c$ in prediction \emph{or} in ground truth}},
    \label{eqn:iou}
\end{equation}
or, equivalently, $\frac{f}{2-f}$, where $f$ is the F1 score for class $c$.)
Notice that while IoU for the \textbf{no change} class is not computed, predictions of change for pixels with no change still contribute to the denominator in (\ref{eqn:iou}) and decrease the score. The evaluation server outputs the IoU for each class upon submission.

For the test phase, a different set of 50 tiles and regions of interest is selected. The submission and evaluation formats remain the same.

\begin{table*}[t]
    \centering
    \begin{tabular}{p{0.8in}p{0.8in}p{0.8in}p{0.8in}p{0.8in}}
    \toprule
        Class & water & tree canopy & low vegetation & impervious \\\midrule
        loss & 0.70\% & 8.12\% & 10.20\% & 1.33\% \\
        gain & 0.86\% & 1.05\% & 7.55\% & 10.90\% \\
    \bottomrule
    \end{tabular}
    \caption{Class statistics for the validation set. The total number of pixels in regions of interest is 13,496,574, or about 1.7\% of the total area of the validation tiles; 59.29\% are labeled as \textbf{no change}.}
    \label{tab:class_stats}
\end{table*}

\section{Methods}

\subsection{Baselines}

We study four baseline algorithms, each of which takes as input the 2013 and 2017 NAIP layers and the 2013 and 2016 NLCD layers.

\begin{enumerate}[(1)]
\item \textbf{NLCD diff:} We use solely the NLCD layers to assign each pixel a high-resolution class according to the second column of Table~\ref{tab:nlcd_to_target}; the \textbf{Barren Land} and \textbf{Developed, Low Intensity} classes are assigned \textbf{impervious} and the \textbf{Developed, Open Space} class is assigned \textbf{low vegetation}. This results in two (coarse) land cover maps in the four target classes. The map derived from NLCD 2013 is used as the prediction for 2013; the map derived from NLCD 2016 is used as the prediction for 2017. Land cover change is computed from these two maps.
\item We train small fully convolutional neural networks (5 layers of 64-filter $3\times3$ convolutions with ReLU activation and a logistic regression layer; 151k parameters) to predict the NLCD labels from NAIP imagery. Two FCNs are trained: one for NAIP 2013, targeting NLCD 2013, and one for NAIP 2017, targeting NLCD 2016.

We found the following scheme to work best for creating target labels from the (probabilistic) FCN predictions: the predicted probabilities (at each pixel) of the classes \textbf{Developed, Open Space}, \textbf{Developed, Low Intensity}, and \textbf{Barren Land}, which do not have a dominant high-resolution target class, are set to 0, and the probabilities are renormalized. The remaining classes are then mapped to the four target labels according to the second column of Table~\ref{tab:nlcd_to_target}: the output probability of each target label is the sum of the probabilities of all NLCD classes mapping to it. The target class with highest probability is then taken as the prediction. The land cover change is computed from the two resulting prediction layers.
\begin{enumerate}[(a)]
\item \textbf{FCN / tile: } Two separate FCNs are trained for each of the validation tiles. Overfitting models with a small receptive field to the coarse NLCD labels and evaluating them results in a super-resolution of these labels: the FCNs cannot detect the 30m block structure in the NLCD labels they were trained to predict, so colors and textures are mapped to the classes of the low-resolution blocks in which they are most likely to appear.
\item \textbf{FCN / all: } The same as (a), but a single FCN is trained on the entire dataset. This may give an advantage due to the much larger training data offering more potential for generalization. However, it may also hurt predictions if pixels with similar appearance tend to belong to different classes in distant parts of the study area.
\end{enumerate}
\item \textbf{U-Net / all: } The same as (2b), but with a model of the U-Net family \cite{unet}, specifically, a U-Net with a ResNet-18 \cite{resnet} encoder structure where the first 3 blocks in the ResNet are used in the downsampling path, and convolutions with 128, 64, and 64 filters are used in the corresponding upsampling layers (1.2m total parameters). We use an implementation from the PyTorch Segmentation Models library\footnote{\url{https://github.com/qubvel/segmentation_models.pytorch}}; 
see the accompanying GitHub repo for details.
\end{enumerate}

The IoU scores for each of these models are shown in Table~\ref{tab:baselines}. Some examples of their predictions appear in Figure~\ref{fig:baseline_examples}.

\begin{table*}[t]
    \centering
    \begin{tabular}{lccccccccc}
    \toprule
        Algorithm & $-$W & $-$TC & $-$LV & $-$I & $+$W & $+$TC & $+$LV & $+$I & avg. \\\midrule
         NLCD diff & 0.148 & 0.167 & 0.282 & 0.014 & 0.031 & 0.001 & 0.106 & 0.362 & \bf 0.139 \\
         FCN / tile & 0.641 & 0.436 & 0.407 & 0.091 & 0.255 & 0.073 & 0.302 & 0.528 & \bf 0.342  \\
         FCN / all & 0.584 & 0.644 & 0.601 & 0.400 & 0.292 & 0.181 & 0.607 & 0.716 & \bf 0.503  \\
         U-Net / all & 0.325 & 0.484 & 0.476 & 0.307 & 0.237 & 0.205 & 0.342 & 0.517 & \bf 0.361  \\
    \bottomrule
    \end{tabular}
    \caption{Baseline IOUs. $-$C and $+$C denote loss and gain of class C, respectively.}
    \label{tab:baselines}
\end{table*}

\subsection{Discussion}

\begin{figure}[t]
    \centering
    \includegraphics[width=0.8\textwidth]{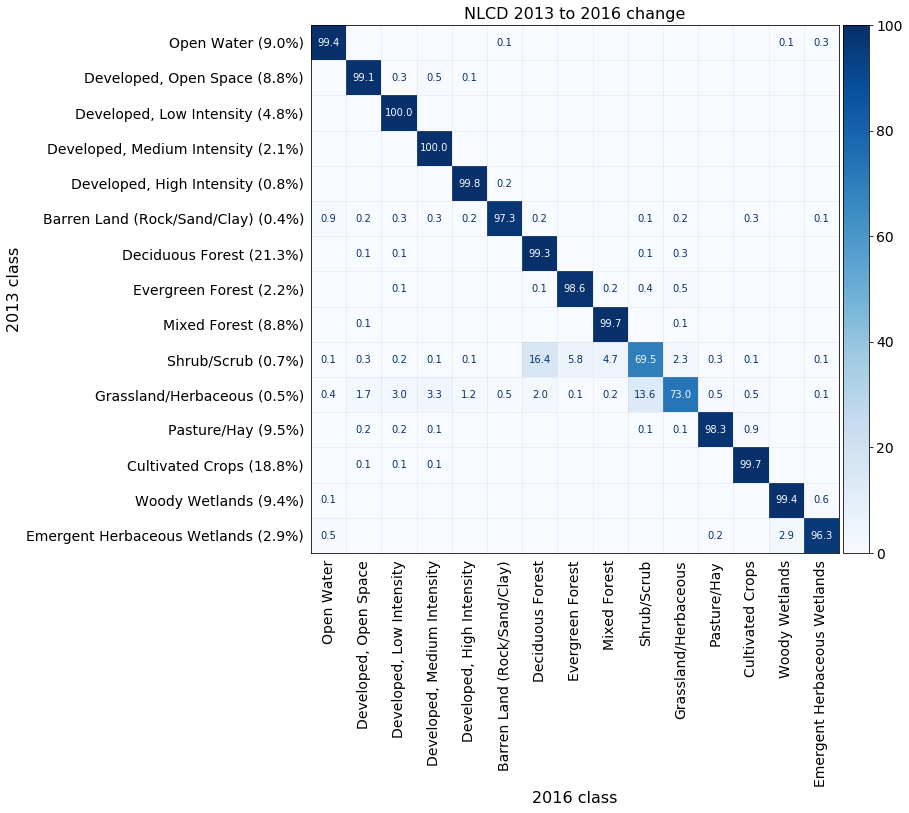}\\
    \includegraphics[width=0.8\textwidth]{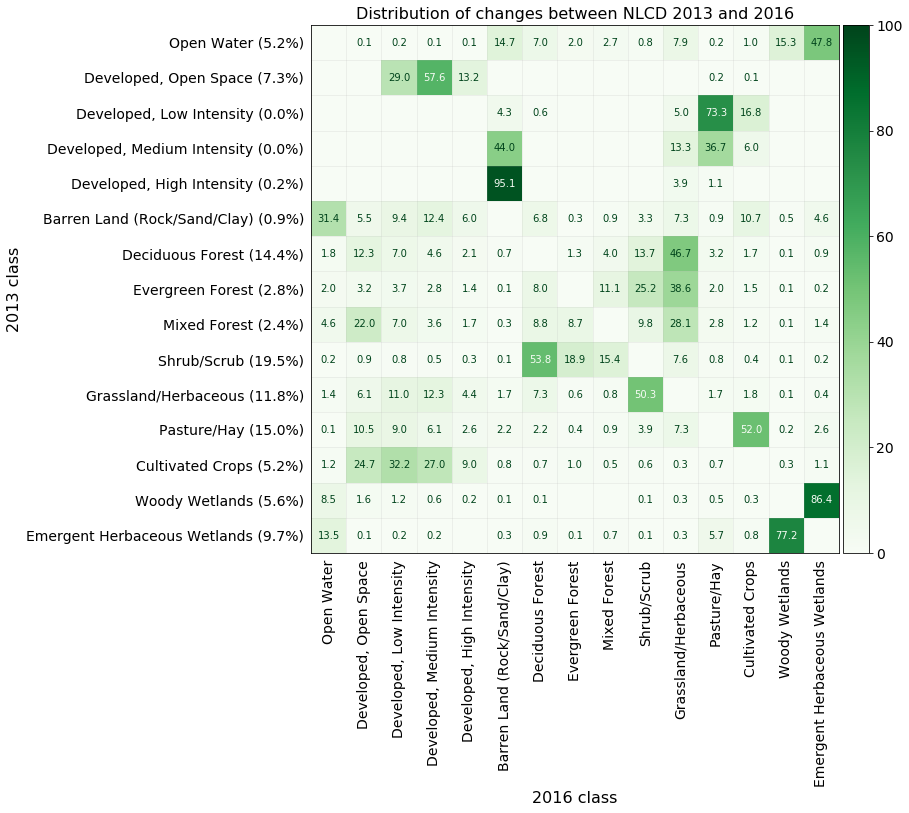}
    \caption{Distribution of changes between 2013 and 2016 NLCD layers. \textbf{Above: } Distribution of 2013 classes (left axis) and distribution of 2016 classes for pixels of each 2013 class. \textbf{Below: } The same, restricted to the set of pixels whose labels differ between 2013 and 2016.}
    \label{fig:nlcd_change}
\end{figure}

From Table~\ref{tab:baselines}, we see that NLCD difference alone is a poor predictor of land cover change. There may be several reasons for this. First, this algorithm gives only coarse change predictions, in 30m blocks, that miss fine details; many changes are in small, thin regions (such as new houses or roads). Second, the 2013 and 2016 NLCD layers were created in a joint pipeline in which certain kinds of change involving the \textbf{Developed} classes were explicitly forbidden \cite{nlcd} (see Figure~\ref{fig:nlcd_change}). Third, the NLCD difference simply misses some changes (e.g., in the lower right corner of Figure~\ref{fig:data_examples}), either because of errors made by the algorithm that produced the NLCD labels, or because NLCD and NAIP represent different time points. 

\begin{figure}[t]
    \centering
    \begin{tabularx}{0.9\textwidth}{m{1cm} l}
    (a) & \includegraphics[width=0.7\textwidth,align=c,trim=4 4 4 0,clip]{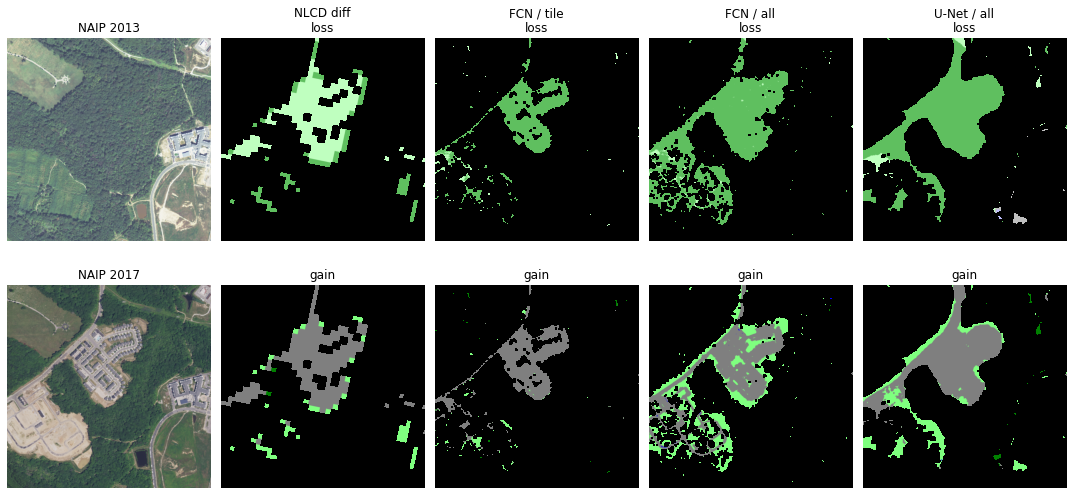} \\ \midrule
    (b) & \includegraphics[width=0.7\textwidth,align=c,trim=4 4 4 22,clip]{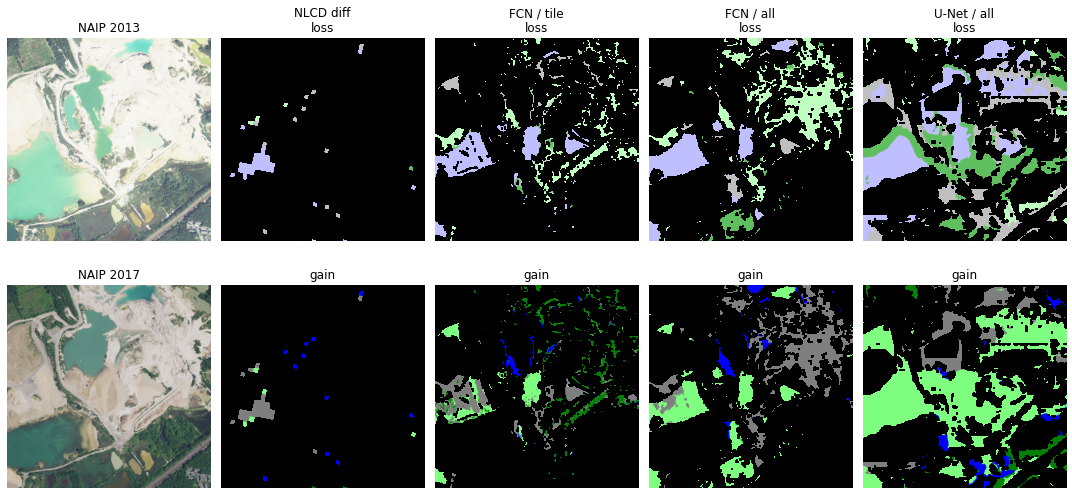} \\ \midrule
    (c) & \includegraphics[width=0.7\textwidth,align=c,trim=4 4 4 22,clip]{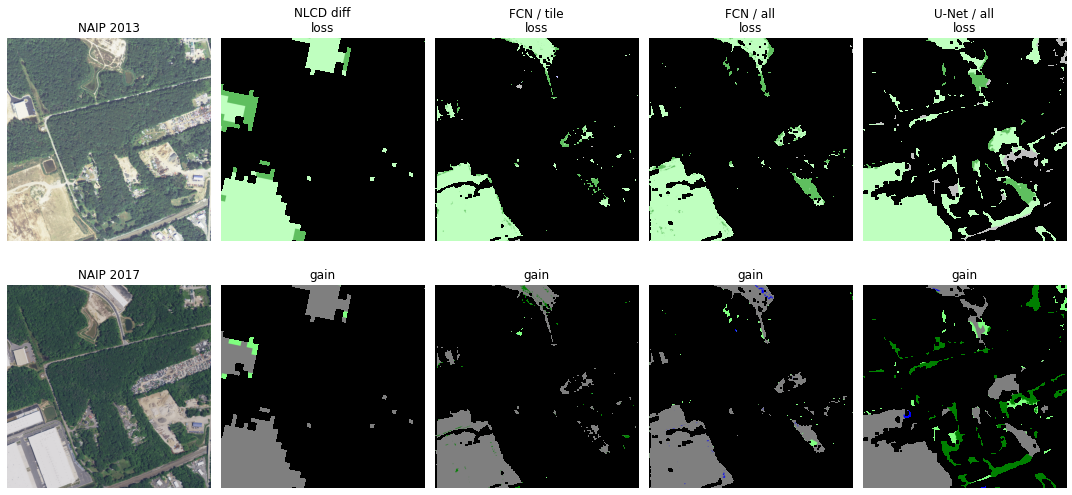} \\ \midrule
    (d) & \includegraphics[width=0.7\textwidth,align=c,trim=4 4 4 22,clip]{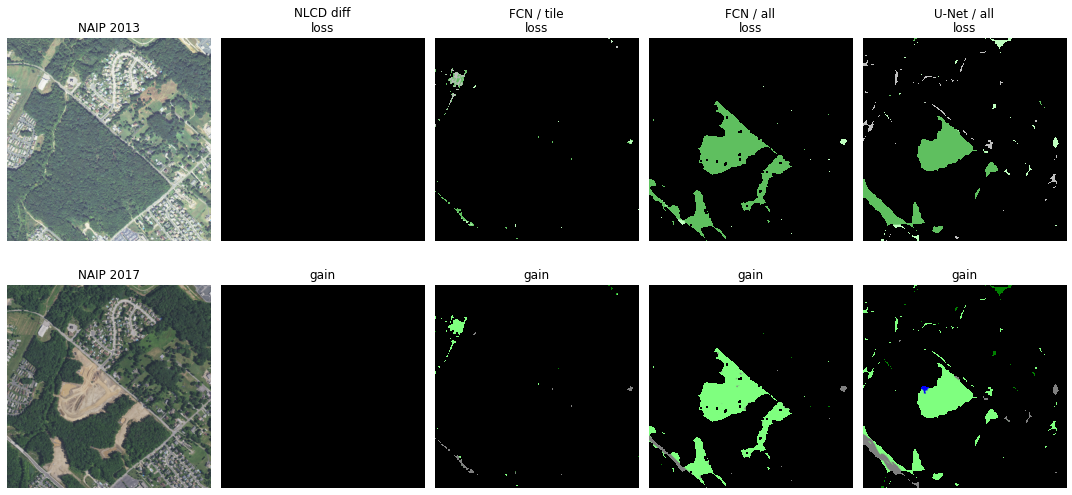} \\ \midrule
    (e) & \includegraphics[width=0.7\textwidth,align=c,trim=4 4 4 22,clip]{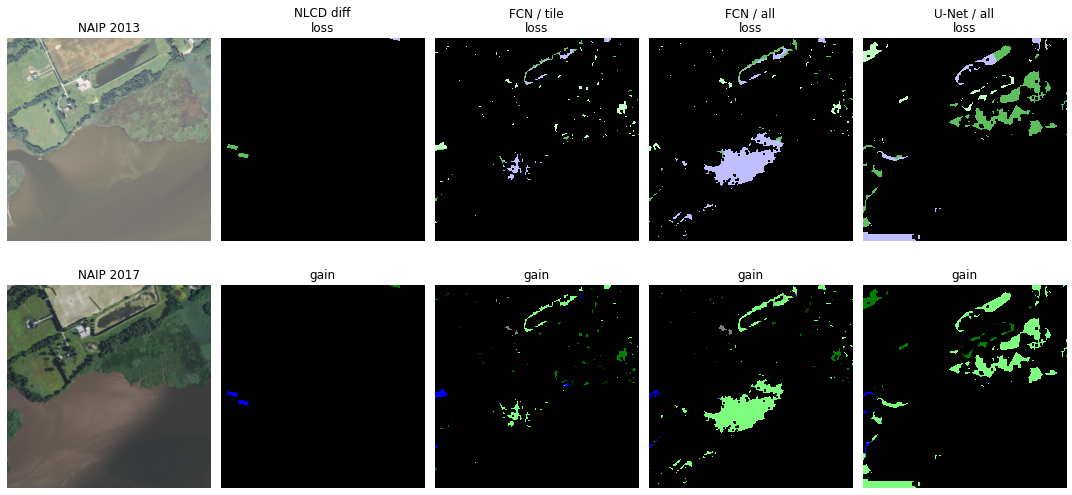}
    \end{tabularx}%
    \caption{Examples of land cover change predictions by baseline models.}
    \label{fig:baseline_examples}
\end{figure}

Interestingly, fitting FCNs to predict the coarse NLCD labels is a strong baseline and an efficient way to super-resolve the NLCD labels to 1m resolution.
In some cases, the FCN trained on a single tile simply ignores changes in areas that have no change in NLCD: it has overfit to the local correspondences of (incorrect) NLCD labels and textures that appear. For example, in Figure~\ref{fig:baseline_examples}(d), it predicts the construction area as \textbf{tree canopy} -- the label suggested by both 2013 and 2016 NLCD labels. However, in other cases, the FCN trained on one tile correctly classifies areas that appear ambiguous to other models: in Figure~\ref{fig:baseline_examples}(e), the silty part of the body of water is predicted correctly as \textbf{water} in both years, as suggested by the \textbf{Open Water} label in both NLCD layers, and thus shown as \textbf{no change}. However, to the FCN trained on all tiles, the local information that this shade of water is indeed \textbf{water} is trumped by evidence from other tiles that this color and texture tends to be \textbf{low vegetation}, resulting in an incorrect prediction.

Differences between class appearances across geographic regions motivated prior work on alternative loss functions for weakly supervised segmentation (label super-resolution) \cite{lsr}; it has also been found that tile-by-tile algorithms sometimes perform better than those trained on a large dataset \cite{epitomes} -- in segmentation problems with wide variation of appearances over space, it is sometimes the case that ``less is more''. We believe that methods that combine global models with tile-by-tile analysis warrant further study.

We also attempted to train models targeting \emph{high-resolution} labels from the Chesapeake Conservancy land cover dataset \cite{chesapeake}, which use a class scheme slightly different from ours. While the use of these labels is not allowed in the DFC-MSD, as they are not available outside of a small geographic region, we found that they do not improve the predictions: surprisingly, models trained on these high-resolution labels over the entire study area perform no better than those trained on coarse NLCD labels.

\begin{figure}[t]
    \centering
    \includegraphics[width=0.9\textwidth]{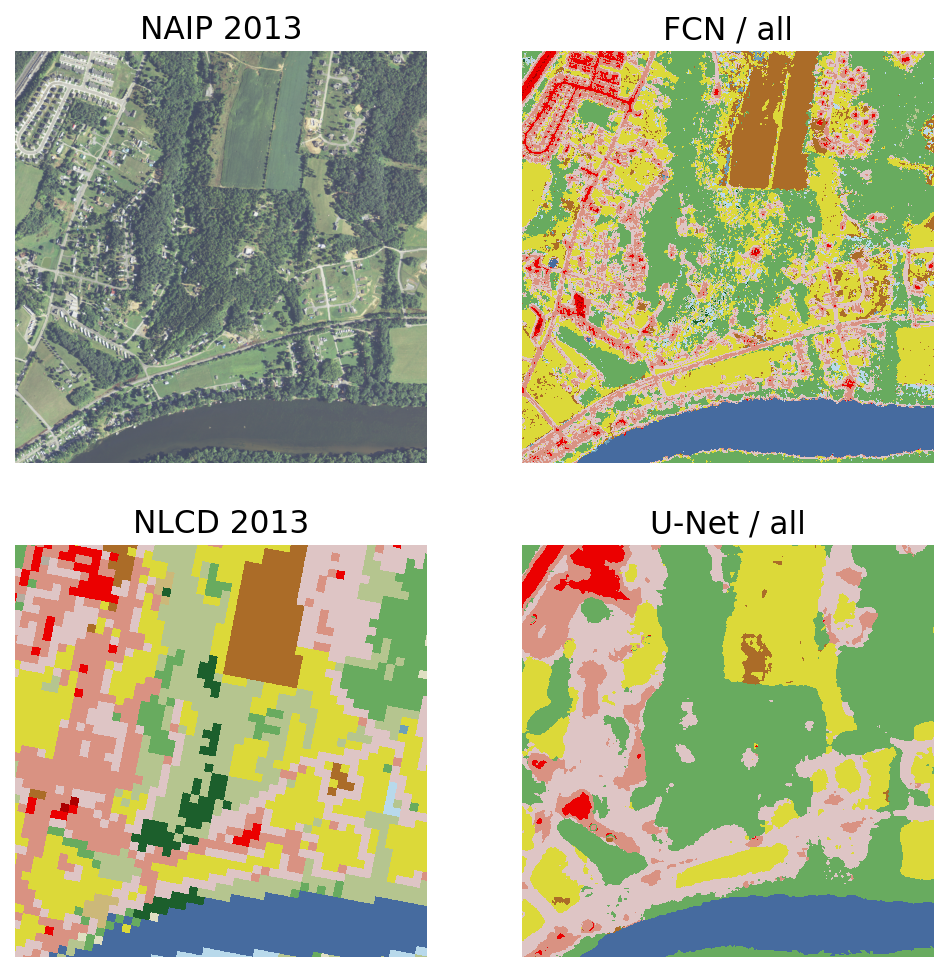}
    \caption{Outputs of a FCN and a U-Net trained on the entire dataset to predict NLCD 2013 classes (bottom left) from NAIP 2013 imagery (top left).}
    \label{fig:fcn_unet}
\end{figure}

The last two rows of Table~\ref{tab:baselines} suggest that sometimes ``less is more'' when it comes to neural architectures as well. The U-Net has about $80\times$ more parameters than the FCN. Figure~\ref{fig:fcn_unet} shows the raw predictions of NLCD labels from the two models, both trained on the entire dataset. The FCN, which has a small ($5\times5$) receptive field, is unable to learn that NLCD labels come in $30\times30$ blocks, while the U-Net has a large receptive field and learns to make blurry predictions. (In fact, the winning weak supervision entry in the 2020 Data Fusion Contest \cite{contest2020} used very simple texture/color mixture models to super-resolve MODIS labels, followed by smoothing with a small FCN.)

We also found that targeting the same year of labels for both 2013 and 2017 networks does not change the results much, as the models primarily learn their textures from the vast area where there is no change anyway. In fact, we should note that training separate models for different time points is a starting point, but we expect that the best approaches will analyze imagery jointly, since most pixels do not change their class from one year to the next. Such situations are often exploited in computer vision, e.g., in background subtraction or co-segmentation. The probabilistic index map model \cite{pim}, for example, learns to assign a distribution over labels to each pixel in an aligned image stack assuming that within each image assignments are consistent, but the palette of assigned colors (or textures) to each index can freely change. 

\section{Conclusion}

In this paper, we have described simple algorithms for land cover change detection and raised questions for future research to answer: How can machine learning models optimally combine noisy, low-resolution labels for detection and classification of change in high-resolution images? How should they strike the balance between local analysis and models that work over a geographically diverse study area? How can low-resolution (Landsat) imagery aid in change detection? The 2021 IEEE GRSS Data Fusion Contest will help to answer these questions, fostering innovation in weakly supervised segmentation and change detection in this important application domain. 

\bibliographystyle{abbrv}
\bibliography{references}

\end{document}